\documentclass[10pt, conference, compsocconf]{IEEEtran}
\ifCLASSINFOpdf
   \usepackage[pdftex]{graphicx}
\else
\fi
\usepackage{amsmath}
\usepackage{amssymb}
\usepackage{multirow}
\usepackage[normalem]{ulem}
\useunder{\uline}{\ul}{}

\usepackage{stfloats}
\begin{document}
%
\title{Pattern-Aware Data Augmentation for LiDAR 3D Object Detection}


\author{\IEEEauthorblockN{Jordan S. K. Hu, Steven L. Waslander}
\IEEEauthorblockA{Institute for Aerospace Studies\\
University of Toronto\\
Toronto, Canada\\
\{jordan.hu, steven.waslander\}@robotics.utias.utoronto.ca}
}



\maketitle

\begin{abstract}
Autonomous driving datasets are often skewed and in particular, lack training data for objects at farther distances from the ego vehicle. The imbalance of data causes a performance degradation as the distance of the detected objects increases. In this paper, we propose pattern-aware ground truth sampling, a  data augmentation technique that downsamples an object's point cloud based on the LiDAR's characteristics. Specifically, we mimic the natural diverging point pattern variation that occurs for objects at depth to simulate samples at farther distances. Thus, the network has more diverse training examples and can generalize to detecting farther objects more effectively. We evaluate against existing data augmentation techniques that use point removal or perturbation methods and find that our method outperforms all of them. Additionally, we propose using equal element AP bins to evaluate the performance of 3D object detectors across distance. We improve the performance of PV-RCNN on the car class by more than 0.7 percent on the KITTI validation split at distances greater than 25 m.
\end{abstract}


%
\IEEEpeerreviewmaketitle

\section{Introduction}
3D object detection has emerged as a prominent challenge on the way to advanced autonomy systems, especially for the application of autonomous driving.  Deep learning techniques have dramatically advanced progress in this domain, however, much remains to be done. One active area of research lies in advancing the capabilities of 3D object detection at long range. Rather than purchasing expensive sensors for more accurate 3D object detection, algorithms can be specifically designed for robust detections in the far range setting as a cheaper alternative. Along with autonomous driving, accurate long range object detection has many other potential applications as well, from intersection traffic monitoring, aerial vehicle security to long-range inspection of infrastructure, for example.

LiDAR is an integral sensor for autonomous driving and many of the state-of-the-art (SOTA) methods rely entirely on this one sensor~\cite{Ge2020, He2020, Lang2019, Shi2020, Shi2019}. LiDARs use laser light to measure distances, creating a 3D point cloud of the scanned environment with high depth accuracy. The captured point cloud can be processed through various methods for accurate 3D object detection.

\begin{figure}[ht!]
\centering
\begin{tabular}{cc}
    \includegraphics[width=0.31\columnwidth]{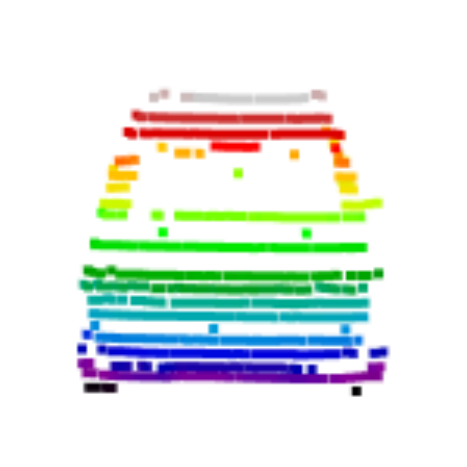}
    &
    \includegraphics[width=0.31\columnwidth]{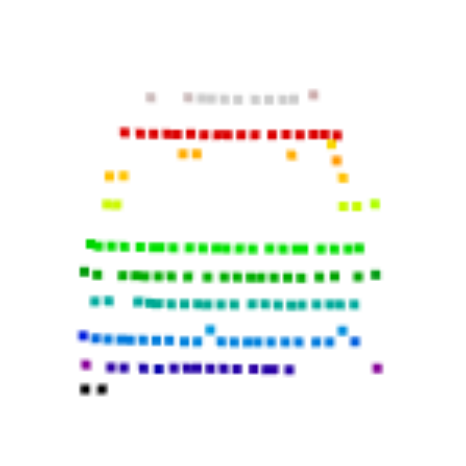}\\
    (a) & (b)
\end{tabular}
\caption{(a) A car point cloud detected at 15 m from the KITTI dataset. (b) A downsampled version of the car point cloud is made by removing every other scan line and removing points in each remaining scan line, simulating an object detected at 30 m.}
\label{fig:sparse-gt}
\end{figure}

\begin{figure}[ht!]
\centering
\begin{tabular}{cc}
    \includegraphics[width=0.45\columnwidth]{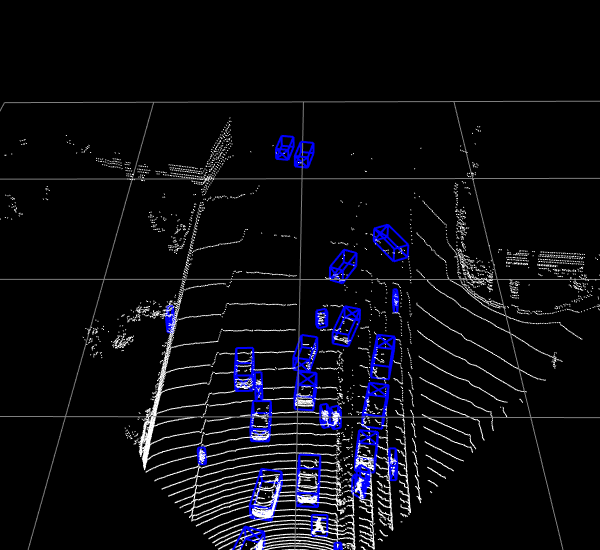} &
    \includegraphics[width=0.45\columnwidth]{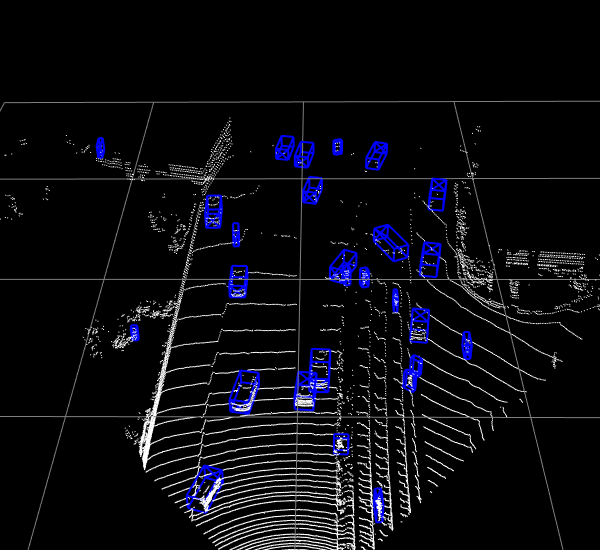}\\
    (a) & (b)
\end{tabular}
\caption{A LiDAR point cloud used for training where the blue boxes indicate the ground truth boxes within the scene. (a) shows the point cloud augmented with regular ground truth sampling whereas (b) uses our pattern-aware ground truth sampling, increasing the variation of ground truth locations.}
\label{fig:sparse-ex}
\end{figure}

However, there is still room for improvement in current LiDAR 3D object detectors, particularly at farther distances. The performance of object detectors tends to degrade as the distance from the LIDAR increases, as shown in Figure \ref{fig:AP-distance}. The problem with detecting objects at far range is two-fold. The first reason stems from the LiDAR itself. LiDARs such as the Velodyne HDL-64E use a series of vertically positioned lasers, creating scan lines across the environment. However, as the distance from the sensor increases, the distance between points increases. This is inherent to the angular resolution of the LiDAR. Thus, the point cloud density on detected objects decreases as we increase the distance from the sensor. The reduction of point cloud data on the detected objects makes it difficult for the algorithm to predict accurate bounding boxes and the performance therefore suffers as a function of range to an object.

\begin{figure*}[ht!]
\centering

\begin{tabular}{ccc}
    \includegraphics[width=0.64\columnwidth]{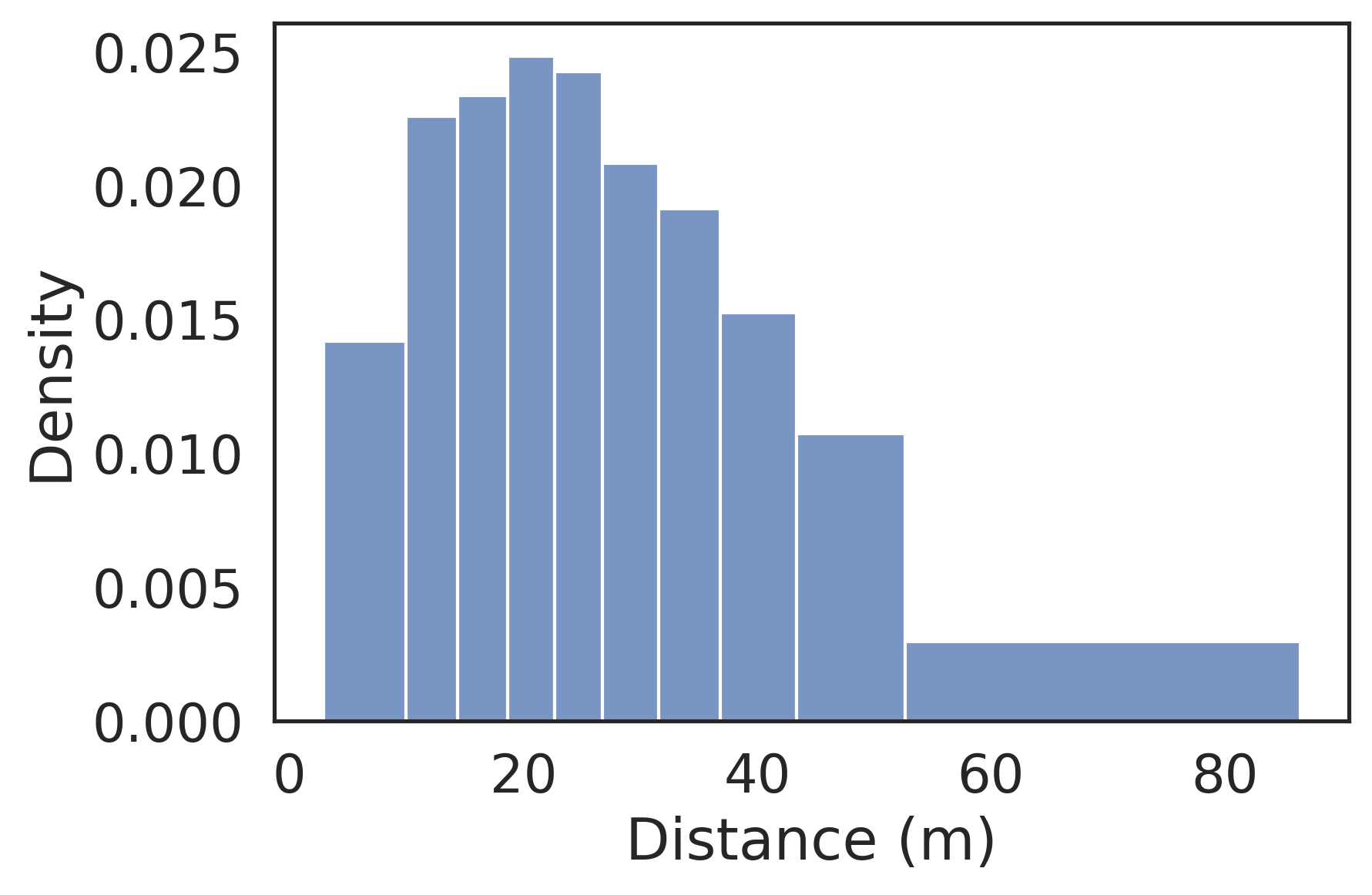} &
    \includegraphics[width=0.64\columnwidth]{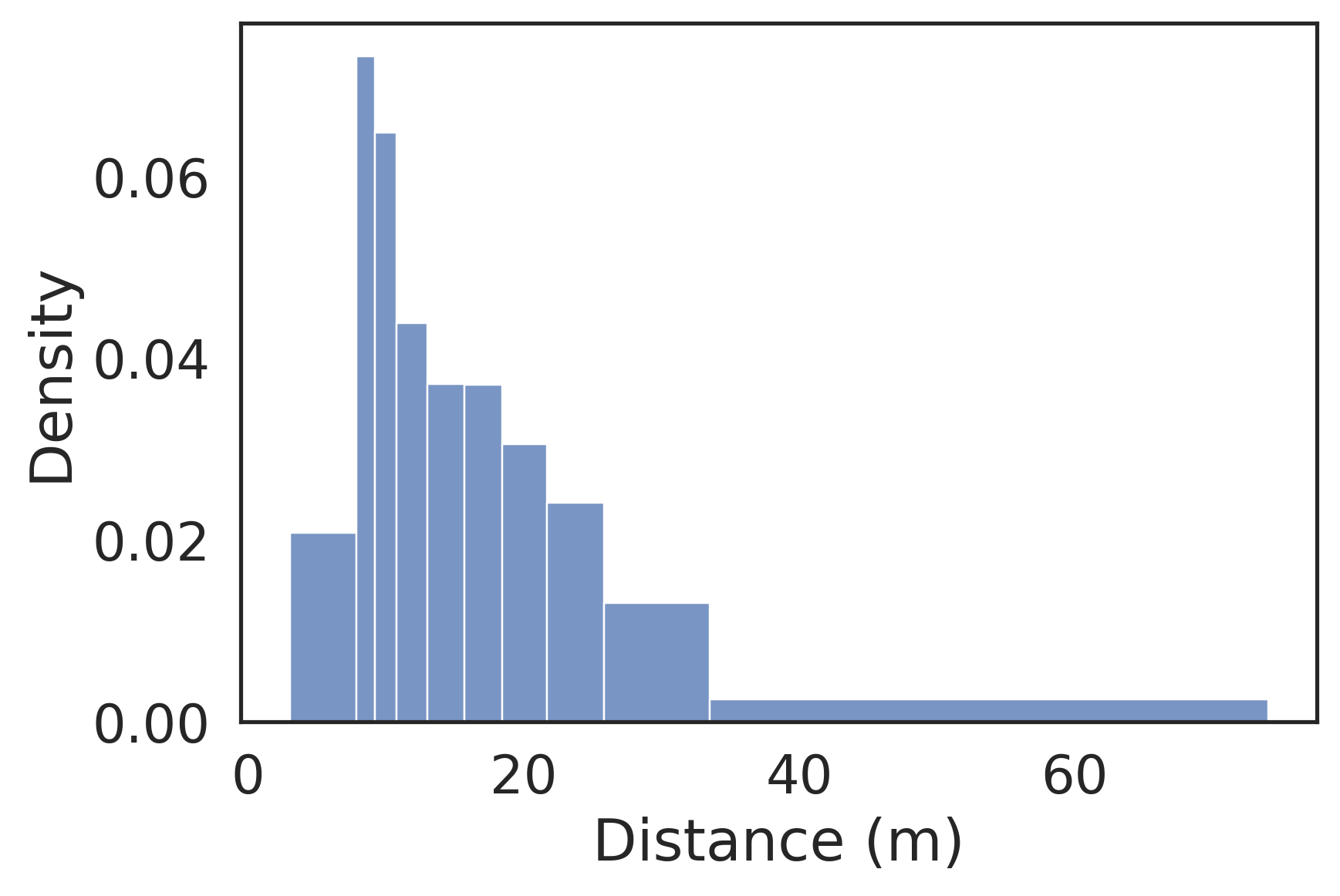} &
    \includegraphics[width=0.64\columnwidth]{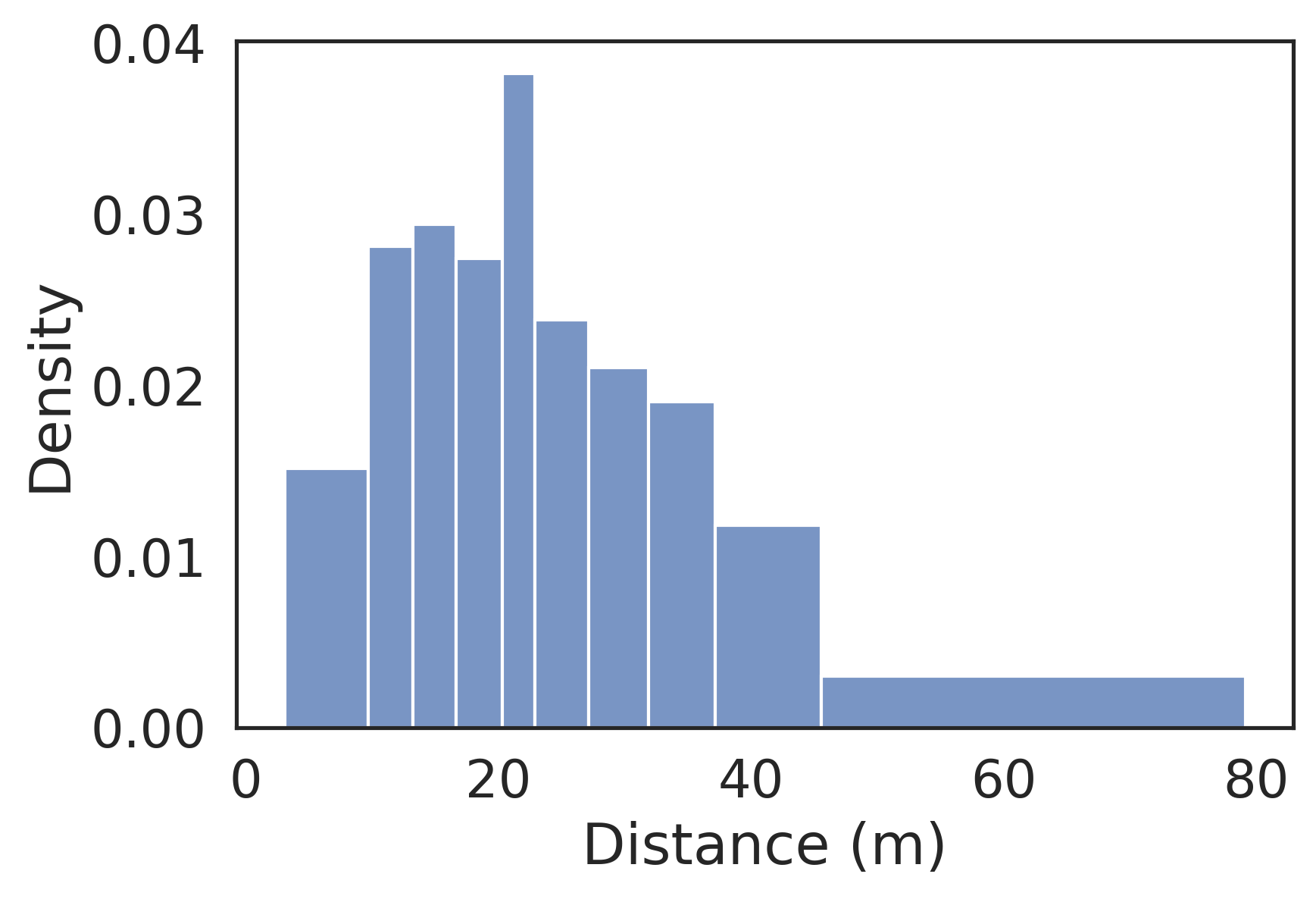}\\
    (a) & (b) & (c)
\end{tabular}
\caption{Histogram density plots showing the distribution of the (a) car, (b) pedestrian, and (c) cyclist classes in the KITTI training dataset. The plots show a skew towards closer objects.}
\label{fig:hist}
\end{figure*}

The second reason is the lack of training data. For example, KITTI \cite{Geiger2012}, a popular dataset used for 3D object detection, is a relatively small dataset since it provides less than 8,000 LiDAR point clouds for training. In comparison, COCO \cite{Lin2014}, an image dataset for 2D object detection, segmentation, and captioning,
contains 330,000 images. SOTA 3D object detection methods typically use data augmentations on the KITTI dataset to artificially increase the small number of training examples. Another problem arises in the imbalance of data across distance. Figure \ref{fig:hist} displays ground truth examples in the KITTI training dataset across distance, showing a skew towards closer objects. Of course, closer objects are more common in scenes and are easier to label, but there is still a bias towards training examples that are closer than 25 m. This results in less training data at long range, causing poorer object detection performance at farther distances.

We propose a solution to resolve the imbalance in training data through a novel pattern-aware ground truth augmentation. By artificially downsampling the point cloud data based on the LiDAR characteristics as shown in Figure \ref{fig:sparse-gt}, we are able to obtain more training examples at farther distances and subsequently increase the performance of detected objects at far ranges.

The benefit of different architectures or data augmentations is usually determined by the average precision (AP) score on the associated dataset. Indeed, AP offers a simple metric to quickly evaluate different algorithms against each other to show the relative value of new contributions within the field. However, the caveat is that it is hard to distinguish the benefit of different methods from a single number. For example, an object detection algorithm can focus on close range examples to boost its overall performance on the benchmark, but is ultimately a poor object detector given its poor performance at farther ranges.

A better method, similar to the Waymo Open Dataset \cite{Sun2020} evaluation metric, is to use distance based AP evaluations. However, naively evaluating samples in equal bin widths has a few issues. As mentioned previously, the data is skewed in KITTI across distance. If the dataset is naively partitioned into equally sized bins (such as 10-20 m, 20-30 m, etc.), there will be an unequal number of samples in each bin. The resulting AP score in each bin therefore is also dependent on the number of samples within said bin. This could result in a poor representation of the object detector's performance in bins with a small number of samples. Rather, we propose to use bins with an equal number of elements and varying size. Thus, the evaluation will not only capture the data evenly, but will also be flexible to any dataset distribution.

To summarize, our contributions are as follows:

\begin{itemize}
    \item We propose a novel method for creating farther distance objects through pattern-aware ground truth sampling
    \item We increase the performance of PV-RCNN by more than 0.7\% at distances greater than 25 m for the car class on the KITTI validation split
    \item We demonstrate the value of using equal element AP bins across distance to capture more insight into the performance of object detectors
    \item We evaluate our approach against other methods that use point removal or perturbation techniques and find that our method outperforms all of them.
\end{itemize}

\section{Related Work}

\subsection{Data Augmentations in LiDAR 3D Object Detection}\label{ssec:related-work-data-aug}
Following Hahner et al.'s \cite{Hahner2020} categorization of different data augmentations, a distinction is made between global and local augmentations. Global augmentations are when the entire point cloud is transformed by the specified augmentation. An example of this would be scaling the entire point cloud by a specified factor. A local augmentation only operates on the points within a ground truth box, such as rotating a bounding box and the associated points. Therefore, local augmentations can take into consideration the local characteristics of the ground truth box, such as the class or the number of points within the box, whereas global augmentations operate on the point cloud as a whole.

VoxelNet \cite{Zhou2018} uses affine transformations for augmenting point clouds in autonomous vehicle datasets and these augmentations are widely used in SOTA models. These include scaling, rotation, and translation of the point clouds. A transformation factor is typically sampled from a random distribution to slightly perturb the point cloud. PIXOR \cite{Yang2018} uses another affine transformation in the form of randomly flipping the point cloud on the vertical axis, similar to the vertical flip augmentation in images. Affine transformations can be performed as global or local point cloud augmentations, but they do not take into consideration the characteristics of the LiDAR when perturbing the points.

SECOND \cite{Yan2018} introduces ground truth augmentation. First, a database that contains all the 3D ground truth boxes is created. During training, the database is sampled to insert additional ground truth samples into the current point cloud sample. If any new ground truth boxes overlap with existing object boxes, they are not included in the point cloud. Ground truth sampling does not alleviate the imbalance in training data because the sampled database has the exact same distribution of the original point cloud data across distances.

Recently, new methods for augmenting point clouds have been proposed. PointMixup \cite{Chen2020} interpolates data between examples through an optimal assignment of the path function between two point clouds. Choi et al. \cite{Choi} introduces PA-AUG, which divides objects into partitions and stochastically applies different augmentations to the segmented portions. Again, these methods treat the point clouds strictly as 3D points in space without considering the sampling pattern of the sensor.

There has also been work on automating the data augmentation parameters throughout training. PointAugment \cite{Li2020} uses an adversarial learning strategy where the augmentation network and the classifier network are jointly trained to create a classifier that is robust to intra-class data variations. Progressive Population Based Augmentation (PPBA) \cite{Cheng2020} uses an evolutionary search algorithm to create a data augmentation scheduling policy.

PPBA also uses additional local data augmentations to increase examples of occlusions and low point density: frustum dropout, frustum noise, and random drop. Both frustum dropout and frustum noise first define a frustum for each ground truth box where the direction of the frustum points from the LiDAR sensor origin to the bounding box. Frustum dropout randomly drops points within the intersection between the ground truth box and the defined frustum with a fixed probability. Frustum noise works similarly but instead perturbs the intersected points with noise sampled from a Gaussian distribution. Random drop simply removes points within a ground truth box based on a fixed probability. These augmentations are similar to pattern-aware ground truth sampling as they attempt to remove or perturb points in the point cloud but they do not consider the sampling pattern of the LiDAR sensor.

\subsection{Sensor Simulation}
There are also methods that attempt to simulate LiDAR data. Tallavajhula et al. \cite{Tallavajhula2018} uses real data to create vegetation and off-road LiDAR simulations. LiDARSim \cite{Manivasagam2020} uses raycasting over 3D scene data and a deep neural network to produce realistic LiDAR point clouds. Other techniques include using CAD models to simulate LiDAR data such as Fang et al. \cite{Fang2020} who combine real world background models and simulated data for improved 3D object detection.

\section{Method}

In this work, pattern-aware ground truth sampling is proposed and evaluated against both common augmentation techniques introduced in SECOND \cite{Yan2018}, as well as additional augmentations outlined in PPBA \cite{Cheng2020}. As mentioned in Section \ref{ssec:related-work-data-aug}, the data augmentations used in PPBA (frustum dropout, frustum noise, and random drop) serve a similar purpose as pattern-aware ground truth sampling by removing or perturbing points. Thus, these augmentation methods serve as an additional baseline for comparison.

Inspired by PseudoLiDAR++'s downsampling method to simulate cheaper 4 and 8 beam LiDARs \cite{You2019}, pattern-aware ground truth sampling downsamples point cloud data based on the LiDAR's characteristics. It builds on top of ground truth sampling \cite{Yan2018} by additionally downsampling ground truth boxes and moving them farther in the scene, simulating farther objects. Pattern-aware ground truth sampling is a local augmentation, operating on each ground truth box.

Pattern-aware ground truth sampling considers the characteristics of the LiDAR sensor itself. As objects move farther away from the LiDAR, each laser point diverges from its neighbours. All lasers are emitted at slightly different angles, allowing for a large coverage area. As the distance from the sensor increases, the scan lines of the LiDAR diverge. For example, the Velodyne HDL-64E has a vertical resolution of 0.4$^\circ$ and a horizontal resolution of 0.08$^\circ$ to 0.35$^\circ$. An object detected at 10 m has a distance of roughly 7 cm between each horizontal scan line whereas an object detected at 40 m has a distance of 28 cm between each scan line. Thus, objects at farther distances will have far fewer scan lines (and points) than closer objects due to the divergence of the lasers from the LiDAR.

\begin{figure}[ht!]
\centering
\includegraphics[width=0.5\columnwidth]{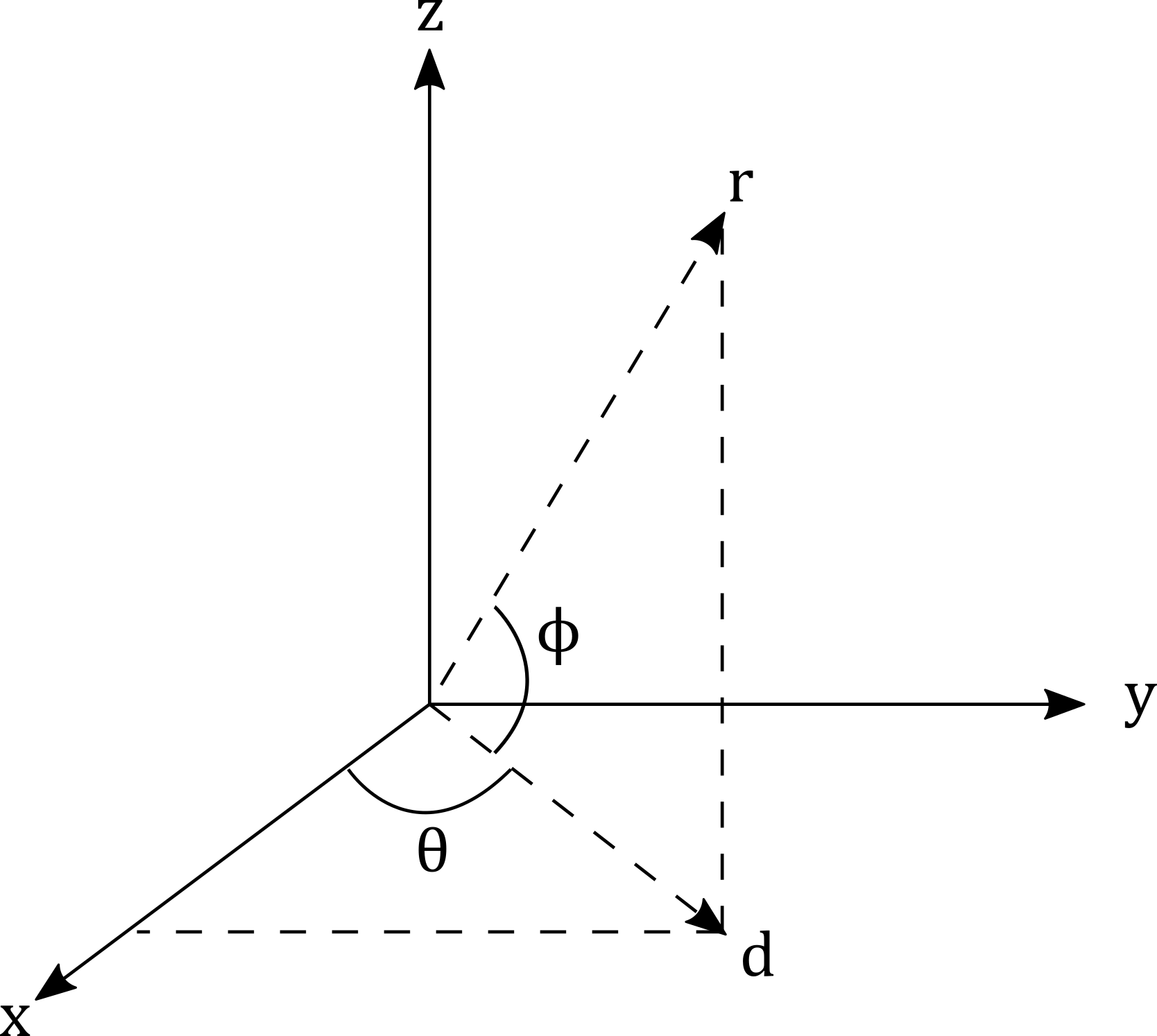}
\caption{Spherical coordinate representation of the point cloud.}
\label{fig:spherical}
\end{figure}

The diverging scan line characteristic of LiDAR data is used to simulate ground truth samples at farther distances from the ego vehicle. A point cloud is defined as a set of 3D points $\{p_i | i=1,\dots,n\}$ where each point $p_i$ has spatial coordinates of $(x_i, y_i, z_i)$. The point cloud can also be represented in its spherical coordinates $(r_i, \theta_i, \phi_i)$ as shown in Figure \ref{fig:spherical}. Thus, $d_i$ and $r_i$ are defined as:
\begin{align}
    d_i = \sqrt{x_i^2 + y_i^2}\\
    r_i = \sqrt{x_i^2 + y_i^2 + z_i^2}
\end{align}
and the azimuthal angle $\theta_i$ and the polar angle $\phi_i$ are calculated as follows:
\begin{align}
    \theta_i = \sin^{-1}\left(\frac{y_i}{d_i}\right)\\
    \phi_i = \sin^{-1}\left(\frac{z_i}{r_i}\right)
\end{align}

\begin{figure*}[ht!]
\centering
\begin{tabular}{ccc}
    \includegraphics[width=0.64\columnwidth]{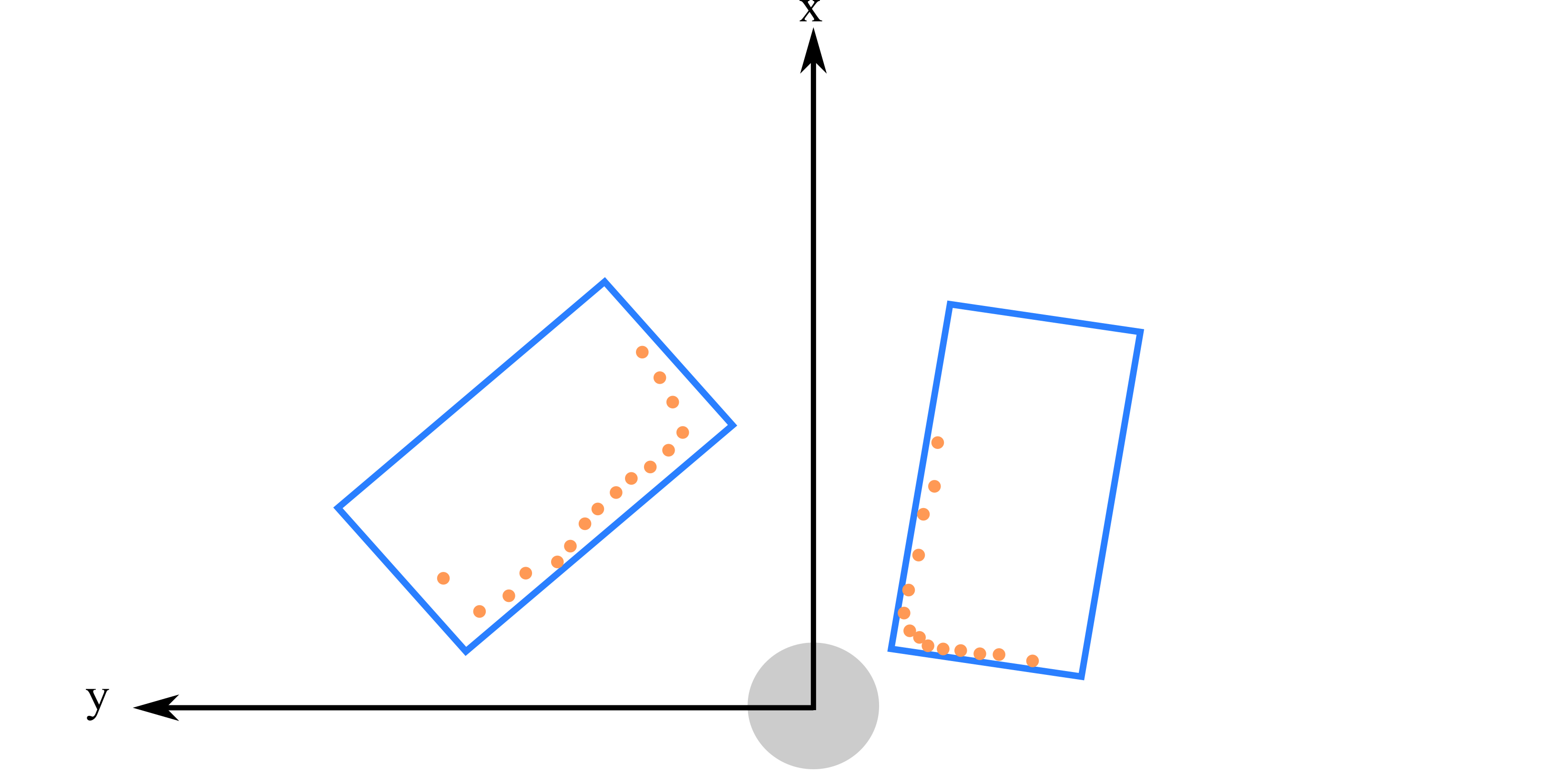} &
    \includegraphics[width=0.64\columnwidth]{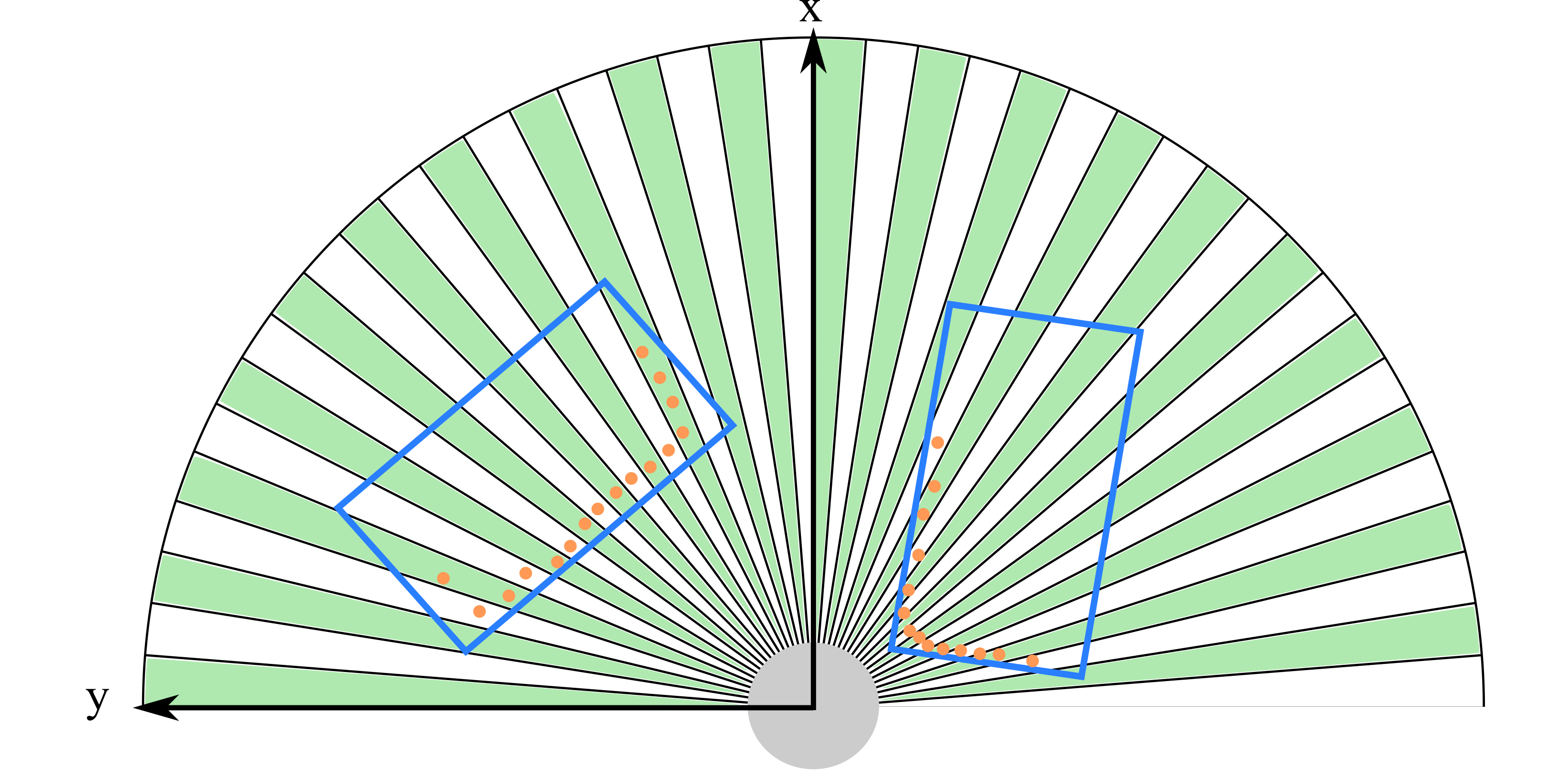} &
    \includegraphics[width=0.64\columnwidth]{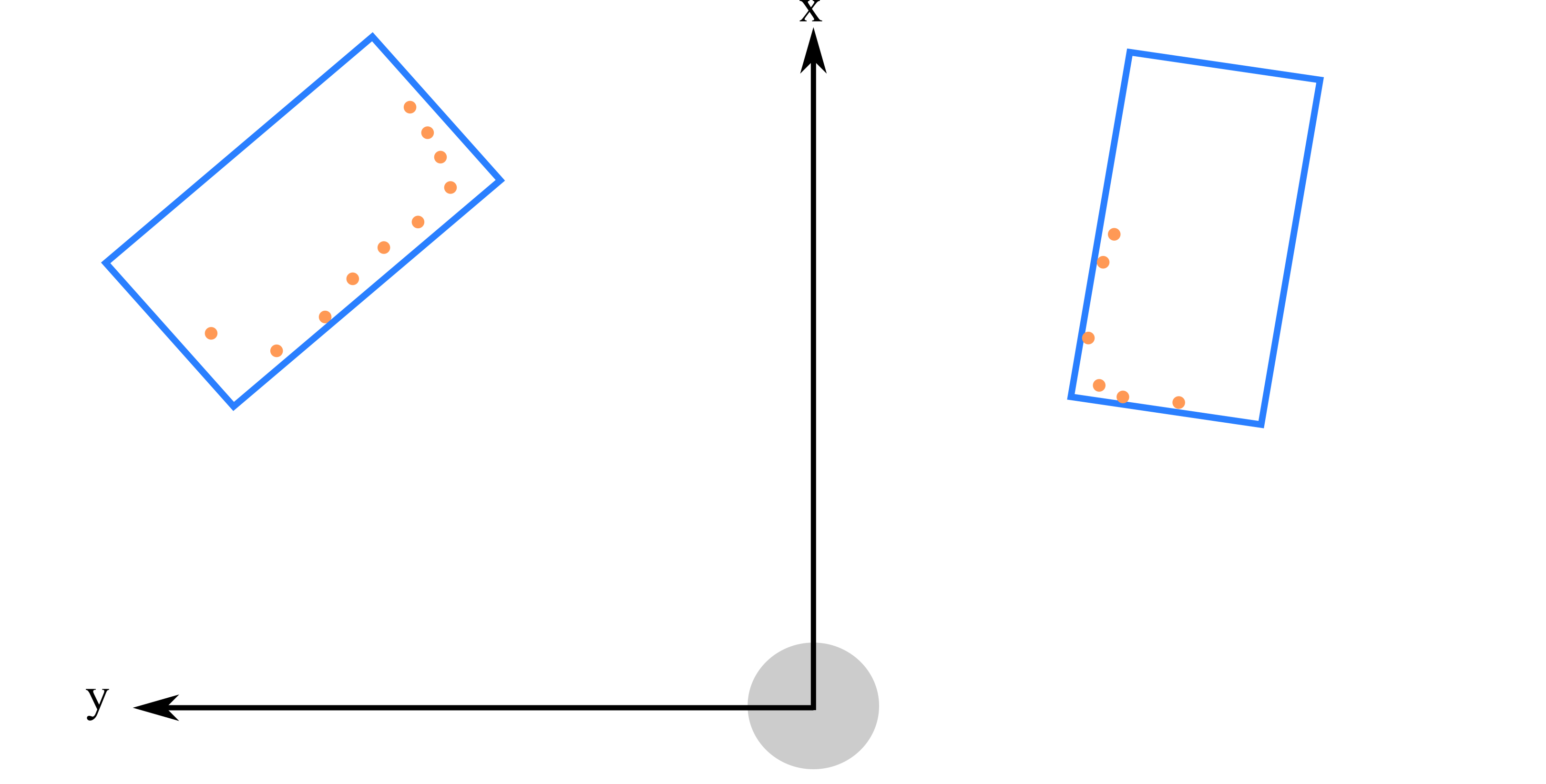}\\
    (a) & (b) & (c)
\end{tabular}
\caption{(a) The original locations of the ground truth boxes (blue) and their point clouds (orange) are inserted into the LiDAR scan. (b) The point cloud is divided across the azimuthal angle $\theta$ into $W$ slices. Then, the points within the ground truth box are downsampled by selecting points in every other slice (green). (c) The pattern-aware ground truth sampled objects now simulate objects detected twice as far as their original locations and are moved farther into the LiDAR scan for additional training examples at farther distances. The same process occurs for the polar angle $\phi$ with $H$ angular divisions.}
\label{fig:bev}
\end{figure*}

Once the point cloud is represented in spherical coordinates, the points can be partitioned across both angular dimensions. Let $W$ be the number of azimuthal angle divisions and $H$ be the number of polar angle divisions. Figure \ref{fig:bev} shows a bird's-eye-view of the angular partitioning of the point cloud across the azimuthal angle $\theta$. Since the point cloud is partitioned into evenly spaced angular divisions, it can be sampled accordingly to simulate LiDAR data captured at different distances. For example, a ground truth bounding box's distance from the LiDAR can be doubled by sampling every other slice, effectively downsampling the point cloud based on the divergence of the LiDAR points. The pattern-aware downsampling method realistically simulates farther objects as shown in Figure \ref{fig:sparse-gt}. Figure \ref{fig:sparse-ex} shows an example of pattern-aware ground truth sampling on a point cloud during training. If the original ground truth sampling is used, objects are often clustered at closer distances. By using pattern-aware ground truth sampling, the locations of objects are more varied, allowing the network to train on a diverse set of object distances.

During training, a portion of the regular ground truth samples are moved further into the scene using the stated pattern-aware ground truth sampling algorithm with a fixed probability. This ensures that there is a more even spread of data examples throughout varying distances, allowing the network to learn from data samples at all depths.

\section{Evaluation}
Evaluating objects at different distances is not new as it is a benchmark used for evaluating 3D object detectors on the Waymo Open Dataset \cite{Sun2020}. The distance analysis is valuable as it gives an understanding of the performance degradation of the object detection method at detecting farther objects. However, existing datasets such as KITTI do not have benchmarks designed specifically for evaluating performance across distance. Naively dividing the data into equally sized bins may cause evaluation issues. Each bin will have a different number of samples, producing unrepresentative evaluations on bins with a small number of samples. Thus, it is important to use equal element bins to ensure a constant number of samples for a meaningful AP in each bin.

Evaluating using equal element AP bins is simple: first a histogram of the data distribution within the dataset is created with a chosen number of bins where each bin contains the same number of samples. For example, if there are 10 histogram bins, each bin contains a 10th of the dataset, varying the bin widths according to the spread of the data. The height of the histogram is normalized such that each bin has the same area. Thus, wide bins will have a lower height, indicating a spread in the data within that region of the histogram, while narrow bins will be taller, indicating a high concentration of data in that bin interval. Example histograms of the KITTI dataset distribution are shown in Figure \ref{fig:hist}. The resulting bin widths from the histogram group samples together based on distance and said groups are evaluated using AP. Providing additional bins for evaluation shows general trends in the data that are obscured by the coarse overall AP values common in standard benchmarks, while being flexible to the distribution of the dataset.

\begin{table*}[]
\centering
\caption{3D AP performance comparison of data augmentations on PV-RCNN evaluated with the KITTI validation split. Highest and second highest values are bolded and underlined respectively.}
\label{tab:overall-AP}
\begin{tabular}{l|ccc|ccc|ccc}
\hline
\multicolumn{1}{c|}{\multirow{2}{*}{Augmentations}} & \multicolumn{3}{c|}{Car (IoU = 0.7)}                      & \multicolumn{3}{c|}{Pedestrian (IoU = 0.5)}      & \multicolumn{3}{c}{Cyclist (IoU = 0.5)}          \\
\multicolumn{1}{c|}{}                               & Easy                    & Moderate       & Hard           & Easy           & Moderate       & Hard           & Easy           & Moderate       & Hard           \\ \hline
Default                                             & 91.83                   & 84.50          & 82.40          & {\ul 65.47}    & 58.05          & 53.07          & 90.31          & 71.74          & 67.84          \\ \hline
Frustum Dropout                                     & 91.75                   & 84.16          & 82.45          & 64.85          & 57.47          & 52.82          & {\ul 90.68}    & \textbf{72.19} & 67.71          \\
Frustum Noise                                       & 91.78                   & 84.28          & {\ul 82.47}    & 64.40          & 57.45          & 52.75          & 89.10          & 71.16          & 67.45          \\
Random Drop                                          & 91.87                   & 83.55          & 82.35          & 65.22          & 57.66          & 52.91          & 89.54          & 71.47          & 67.61          \\
Frustum Dropout + Noise                             & {\ul 91.90}             & {\ul 84.56}    & {\ul 82.47}    & 65.28          & {\ul 58.06}    & {\ul 53.35}    & \textbf{91.13} & {\ul 72.13}    & \textbf{68.14} \\ \hline
Pattern-Aware GT (Ours)                             & \textit{\textbf{92.13}} & \textbf{84.79} & \textbf{82.56} & \textbf{65.99} & \textbf{58.57} & \textbf{53.66} & 90.38          & 72.03          & {\ul 67.96}    \\ \hline
\textit{Improvement}                                         & \textit{+0.3}           & \textit{+0.29} & \textit{+0.16} & \textit{+0.52} & \textit{+0.51} & \textit{+0.59} & \textit{+0.06} & \textit{+0.3}  & \textit{+0.12} \\ \hline
\end{tabular}
\end{table*}

\section{Experiments}
\label{sec:experiments}
All models are trained and evaluated on the KITTI dataset. Training is primarily done using PV-RCNN \cite{Shi2020}. Following PV-RCNN, we use the same image split of 3712 training and 3769 validation examples. Each training configuration is run 5 times and averaged. PV-RCNN is trained with the default training parameters and uses ground truth sampling, random global flip, random global rotation, and random global scaling. This serves as a baseline, and all other models use the additional augmentations that are specified.

 Since the LiDAR used in the KITTI dataset is the Velodyne HDL-64E, 512 azimuth and 64 polar channels are used for the number of angular divisions and downsampled to 32 and 256 channels respectively to double the distance of the original objects' location. The minimum number of points after downsampling the objects through pattern-aware ground truth sampling is 5, 200, and 200 for the car, pedestrian, and cyclist class, respectively.  The range of downsampled objects is limited to 20-70 m from the LiDAR. These point and range limits are used to ensure that objects are not downsampled to too few points. Pattern-aware ground truth sampling is performed on the samples with a probability of 40\%.
 
 Alongside the standard KITTI AP benchmark evaluations, the data augmentation configurations are tested using equal element AP bins to evaluate performance across distance. The bin widths are calculated by splitting the validation ground truth samples into 10 equal element bins. AP is calculated using the predicted and ground truth boxes that fall within the specified bin width. Unlike the standard KITTI benchmark, the distance evaluation also includes ground truth examples outside the KITTI hard difficulty \cite{Geiger2012} (Minimum bounding box height: 25 pixels, maximum occlusion level: difficult to see, maximum truncation: 50\%). This is done to evaluate the object detection performance at far distances which are normally excluded in the standard KITTI difficulty levels.

\section{Results}
\subsection{Overall AP}
Table \ref{tab:overall-AP} shows the 3D AP results calculated using 40 recall positions for each augmentation. Using pattern-aware ground truth sampling, PV-RCNN obtains a slight overall AP improvement on the KITTI validation split. A good mix of the original and downsampled ground truth augmentations is necessary for improving performance of the object detector. In fact, a 40\% probability of pattern-aware downsampling the ground truth augmentations provides a good balance of data examples at various distances. A higher percentage results in a smaller increase in performance as the data overrepresents training examples at far range. 

Pattern-aware ground truth sampling is also used on the cyclist and pedestrian classes but it is necessary to set the minumum number of points after downsampling to 200 points. A lower number results in high variation in performance on the two classes. One reason is because of the extreme skewness of data. Figure \ref{fig:hist} (b) and (c) show the imbalance in pedestrian and cyclist classes is extreme compared to the car class. Thus, the distribution of data for these classes can drastically change when using pattern-aware ground truth sampling, causing fluctuations in resulting performance. Another reason is because the pedestrian and cyclist classes are less structured than the car class. Unlike downsampling a rigid object, pedestrian and cyclist point clouds can have various spatial configurations, which produces variable results from pattern-aware ground truth sampling if downsampled to a low number of points.

The resulting performance for the PPBA augmentations are mixed. Frustum dropout provides slight increases in the easy and moderate cyclist class but suffers degradations in the pedestrian class. Both frustum noise and random drop suffer degradations in the cyclist and pedestrian class. All PPBA methods perform slightly worse on the car class overall with slight increases in the hard category.

Performance of the PPBA augmentations is improved by coupling frustum dropout and noise together. There are slight increases in most of the categories except in the easy pedestrian class but even the combined augmentations do not perform as well as pattern-aware ground truth sampling.

\begin{table*}[]
\centering
\caption{PV-RCNN 3D AP performance across distance on the KITTI validation split for the car class. Samples are binned and evaluated across distance from the LiDAR. Highest and second highest values are bolded and underlined respectively.}
\label{tab:distance-AP}
\begin{tabular}{l|cccccccccc}
\hline
\multirow{2}{*}{Augmentations} & \multicolumn{10}{c}{Distance (m)}                                                                                                                                       \\ \cline{2-11} 
                               & 0-9            & 9-13           & 13-17          & 17-21          & 21-25          & 25-30          & 30-35          & 35-41          & 41-50          & 50-85          \\ \hline
Default                        & 85.48          & 91.06          & 92.14          & {\ul 90.55}    & {\ul 86.88}    & 82.48          & 74.34          & 59.69          & 40.58          & 13.66          \\ \hline
Frustum Dropout                & 84.39          & 91.24          & 92.56          & 90.47          & 86.70          & 83.17          & 74.90          & {\ul 60.15}    & {\ul 40.59}    & 13.02          \\
Frustum Noise                  & {\ul 87.57}    & {\ul 91.27}    & \textbf{93.71} & 90.40          & 86.82          & 83.00          & 75.00          & 59.37          & \textbf{40.63} & {\ul 13.68}    \\
Drop Point                     & 86.42          & 91.25          & 91.84          & 90.09          & 86.58          & 82.47          & 74.07          & 59.76          & 39.21          & 13.46          \\
Frustum Dropout + Noise        & 84.71          & 91.11          & 92.13          & 90.52          & \textbf{86.92} & {\ul 83.21}    & {\ul 75.58}    & 60.05          & 39.63          & 13.66          \\ \hline
Pattern-Aware GT (Ours)        & \textbf{87.78} & \textbf{91.68} & {\ul 93.13}    & \textbf{90.65} & 86.80          & \textbf{83.80} & \textbf{75.83} & \textbf{60.41} & 39.98          & \textbf{14.63} \\ \hline
\textit{Improvement}           & \textit{+2.30}  & \textit{+0.62} & \textit{+0.99} & \textit{+0.10}  & \textit{-0.08} & \textit{+1.32} & \textit{+1.49} & \textit{+0.72} & \textit{-0.60} & \textit{+0.97} \\ \hline
\end{tabular}
\end{table*}

\subsection{AP Across Distance}
\begin{figure}[ht!]
\centering
\includegraphics[width=0.85\columnwidth]{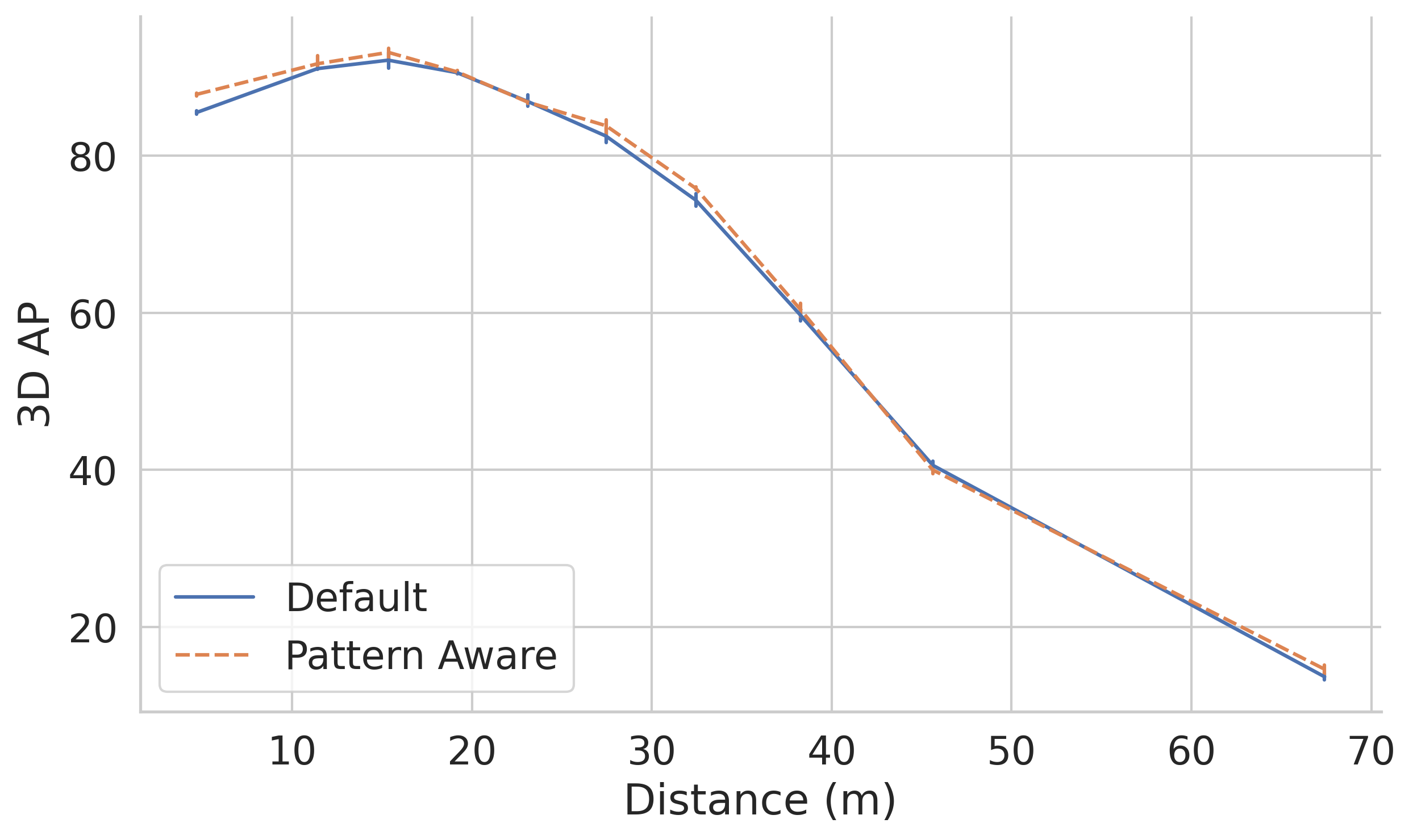}
\caption{PV-RCNN 3D AP performance across distance on the KITTI validation split for the car class. The blue solid line indicates the default training and the dashed orange line is the performance with the addition of pattern-aware ground truth sampling. The error bars indicate the 95\% confidence interval of the 5 averaged trainings.}
\label{fig:AP-distance}
\end{figure}

The performance gain from pattern-aware ground truth sampling is more evident when the model's performance on 3D AP is plotted using equal element AP bins across distance as shown in Figure \ref{fig:AP-distance} and Table \ref{tab:distance-AP}. There is a general increase in AP across distance. Within the 25-41 m and 50-85 m range, there is an increase in of more than 0.7\% with the largest increase of 1.49\% within the 30-35 m range. Pattern-aware ground truth sampling adds more diverse object locations, increasing performance across distance. The caveat, of course, is the degradation in the 21-25 m and 41-50 m range. This is attributed to the shift in the training data distribution compare to the validation data distribution which is further explained in Section \ref{ssec:data-imbalance}.

Another benefit from pattern-aware ground truth sampling is the increase in AP at closer distances, especially in the 0-9 m range. Within this range, many objects are truncated due to the frustum segmented point cloud. Even though pattern-aware ground truth sampling is specifically targeted towards increasing performance at far range, only objects between the 20-70 m range are downsampled. By reducing the training imbalance across distance, there is a larger percentage of training data at close distances and therefore the trained 3D object detector performs better at closer range. Without the analysis of equal element AP bins, these two observations would not have been easy to discern from the coarse standard KITTI difficulties. Using equal element AP bins provides insightful breakdowns of how a 3D object detector is performing.

The PPBA augmentations overall have some improvements across distance, specifically frustum noise and the combination of frustum dropout and frustum noise. However, pattern-aware ground truth sampling performs substantially better overall.

\begin{figure}[ht!]
\centering
\begin{tabular}{cc}
    \includegraphics[width=0.45\columnwidth]{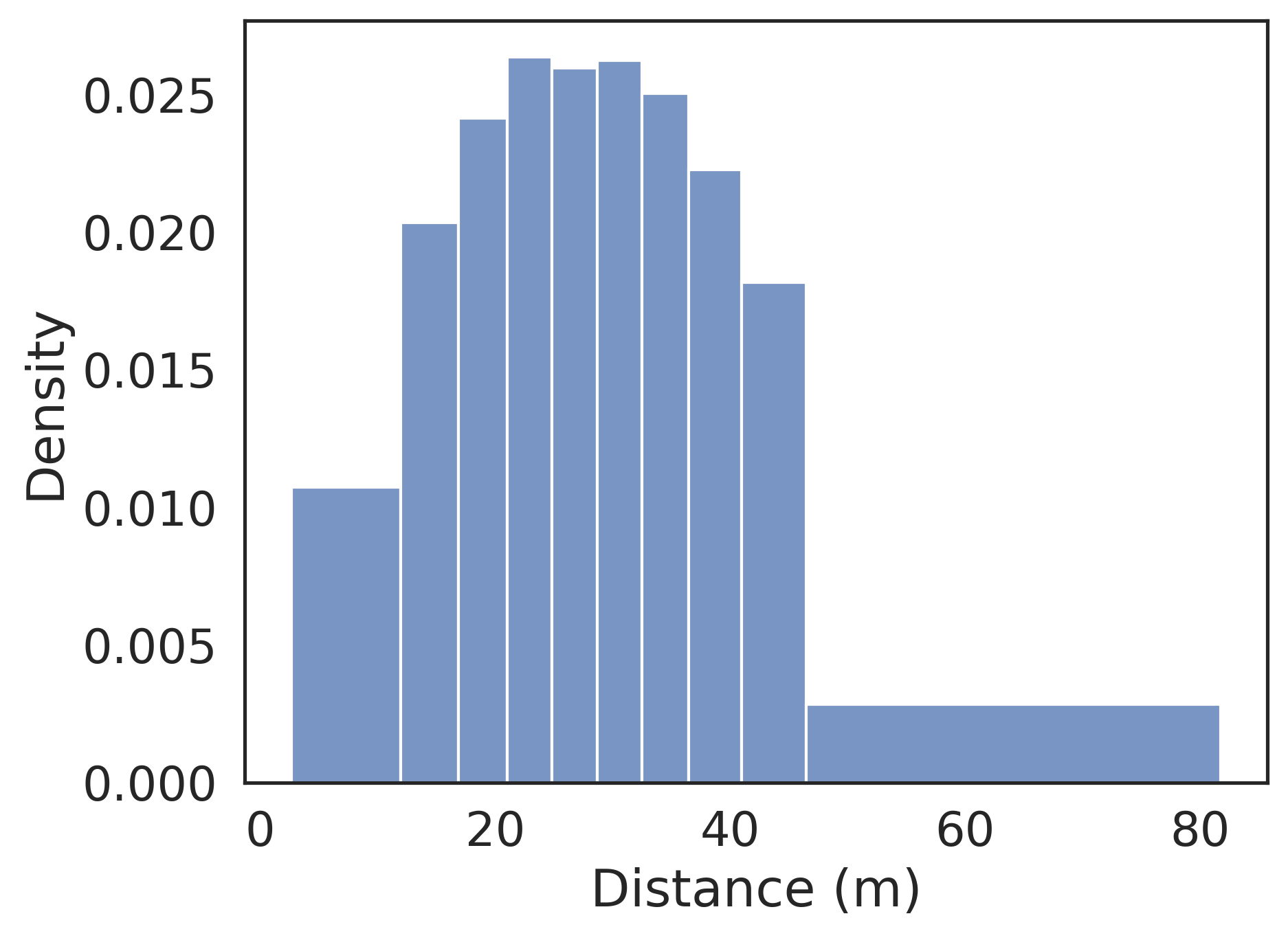} &
    \includegraphics[width=0.45\columnwidth]{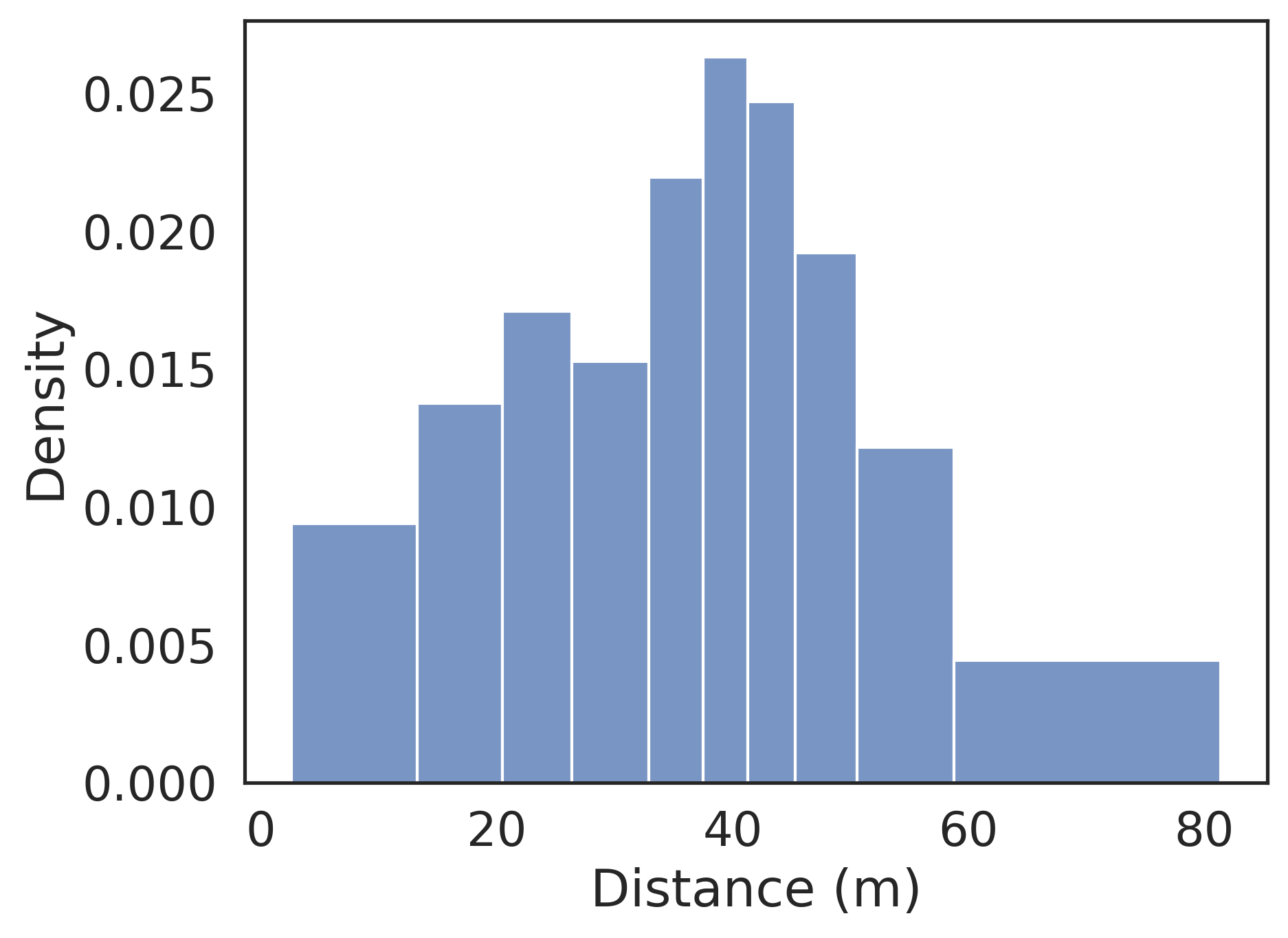}\\
    (a) & (b)
\end{tabular}
\caption{The data distribution of the car class (a) before and (b) after using pattern-aware ground truth sampling on the KITTI dataset. 5000 ground truth objects are sampled and the distribution of data is plotted. Pattern-aware ground truth sampling shifts the dataset towards farther objects, creating a less skewed distribution of data.}
\label{fig:data-imbalance}
\end{figure}

\subsection{Data Imbalance} \label{ssec:data-imbalance}
To understand the implications of pattern-aware ground truth sampling, its effects on the distribution of the car class data are explored. 5000 data points are randomly sampled from the ground truth database with a fixed seed. Figure \ref{fig:data-imbalance} shows the balance of histogram distributions of the car class before and after pattern-aware ground truth sampling across distance from the LiDAR sensor. As expected, the first distribution is similar to Figure \ref{fig:hist} (a) as the regular ground truth sampling uses the same ground truth boxes as the original training set. By adding pattern-aware ground truth sampling, the dataset shifts to a less skewed distribution, alleviating the inherent imbalance in data and providing the network more training examples at farther distances. Ideally, the resulting distribution across distance should be uniform but doing so currently degrades overall performance. Future improvements to pattern-aware ground truth sampling aim to focus on creating a uniform distribution of training examples across distance while retaining SOTA performance.

Even though the distribution is less skewed after pattern-aware ground truth sampling, there is a performance degradation in the 21-25 m and 41-50 m bins. As shown in Figure \ref{fig:data-imbalance}, the highest concentration of each distribution is exactly at the 21-25 m and 41-50 m range respectively. Since the models are trained with a distribution closer to Figure \ref{fig:data-imbalance} (b), the trained models expect less objects in the 20 m range and more objects in the 40 m range, resulting in a performance degradation when evaluating on the validation split which is closer to Figure \ref{fig:data-imbalance} (a)'s distribution. Nonetheless, the poorer performance in these specific bins are outweighed by the overall performance increase of pattern-aware ground truth sampling across distance.

\begin{table*}[]
\centering
\caption{3D AP performance of pattern-aware ground truth sampling on the KITTI validation split on other LiDAR 3D object detectors. Highest values are in bold.}
\label{tab:general-AP}
\begin{tabular}{l|l|rrr|rrr|rrr}
\hline
\multicolumn{1}{c|}{\multirow{2}{*}{Model}} & \multicolumn{1}{c|}{\multirow{2}{*}{Augmentations}} & \multicolumn{3}{c|}{Car (IoU = 0.7)}                                                & \multicolumn{3}{c|}{Pedestrian (IoU = 0.5)}                                         & \multicolumn{3}{c}{Cyclist (IoU = 0.5)}                                            \\
\multicolumn{1}{c|}{}                       & \multicolumn{1}{c|}{}                               & \multicolumn{1}{c}{Easy} & \multicolumn{1}{c}{Moderate} & \multicolumn{1}{c|}{Hard} & \multicolumn{1}{c}{Easy} & \multicolumn{1}{c}{Moderate} & \multicolumn{1}{c|}{Hard} & \multicolumn{1}{c}{Easy} & \multicolumn{1}{c}{Moderate} & \multicolumn{1}{c}{Hard} \\ \hline
PV-RCNN                                     & Default                                             & 91.80                    & 84.50                        & 82.42                     & 64.77                    & 57.20                        & 52.43                     & 91.20                    & 72.20                        & 68.62                    \\
                                            & Pattern-Aware GT                                    & \textbf{92.07}           & \textbf{84.56}               & \textbf{82.48}            & \textbf{65.04}           & \textbf{58.03}               & \textbf{53.23}            & \textbf{91.51}           & \textbf{72.55}               & \textbf{68.64}           \\ \hline
SECOND                                      & Default                                             & 90.26                    & \textbf{81.57}               & 78.67                     & 57.02                    & 51.84                        & 47.38                     & 82.00                    & 65.21                        & 61.35                    \\
                                            & Pattern-Aware GT                                    & \textbf{90.46}           & 81.55                        & \textbf{78.73}            & \textbf{57.47}           & \textbf{52.40}               & \textbf{47.77}            & \textbf{84.08}           & \textbf{66.86}               & \textbf{62.94}           \\ \hline
PointPillars                                & Default                                             & 87.65                    & 78.35                        & 75.53                     & 54.19                    & 48.53                        & 44.27                     & \textbf{83.47}           & \textbf{64.95}               & \textbf{60.70}           \\
                                            & Pattern-Aware GT                                    & \textbf{87.95}           & \textbf{78.69}               & \textbf{75.81}            & \textbf{57.07}           & \textbf{51.02}               & \textbf{46.33}            & 82.07                    & 64.27                        & 60.28                    \\ \hline
PointRCNN                                   & Default                                             & \textbf{90.10}           & \textbf{80.41}               & \textbf{78.00}            & \textbf{64.18}           & \textbf{56.71}               & \textbf{49.86}            & \textbf{91.72}           & \textbf{72.47}               & \textbf{68.18}           \\
                                            & Pattern-Aware GT                                    & 89.97                    & 80.32                        & 77.93                     & 63.19                    & 55.95                        & 48.97                     & 90.24                    & 71.52                        & 66.96                    \\ \hline
\end{tabular}
\end{table*}

\subsection{Generalizability}
Table \ref{tab:general-AP} shows the performance of the augmentations on SECOND \cite{Yan2018}, PointPillars \cite{Lang2019}, and PointRCNN \cite{Shi2019} on the KITTI car validation split. Pattern-aware ground truth sampling provides a general increase to SECOND and PointPillars, but suffers an overall degradation on PointRCNN. A reason for the variable results in performance may be the differences in architecture. Both SECOND and PointPillars use an established grid of anchor boxes for bounding box predictions, allowing them to encode the pattern-aware ground truth samples in a consistent approach. In comparison, PointRCNN uses a PointNet++ backbone for bounding box proposals and may not benefit from pattern-aware ground truth sampling. The use of farthest point sampling is less structured and may not capture the downsampled ground truth objects well. A possible improvement to pattern-aware ground truth sampling for PointRCNN would be to perform pattern-aware ground truth sampling after the initial farthest point sampling points have been chosen to ensure consistent point sampling.

PointPillars has a tradeoff between the pedestrian and cyclist class performance. The benefit of pattern-aware ground truth sampling does outweigh the drawbacks as there is an overall increase of 2\% in the pedestrian class while there is only a small degradation in the cyclist class. Still, pattern-aware ground truth sampling may need slight tuning to ensure optimal performance. In the case of PointPillars, the cyclist class' minimum number of points for pattern-aware ground truth sampling may need to be increased.

\section{Conclusion and Future Work}
This work presents a method for handling the imbalance of LiDAR data in autonomous driving datasets. Our method utilizes LiDAR characteristics by simulating the diverging lasers to create a more diverse set of training examples across distance. By doing so, 3D object detection  methods are able to perform better at detecting objects at farther distances. Compared to other augmentation methods that remove points, pattern-aware ground truth sampling performs better overall. Pattern-aware ground truth sampling is also trained on other LiDAR 3D object detectors to show its applicability to different architectures, providing performance improvements in grid based LiDAR 3D object detectors. Equal element AP bins give key insights on the impact of pattern-aware ground truth augmentations. Although only small increases in performance are shown on the overall 3D AP, the benefit of the method is found at farther distances, particularly at distances greater than 25 m.

Pattern-aware ground truth sampling does not consider reprojecting the removed points when downsampling objects. Realistically, if a car was detected at a farther distance, the removed laser points would hit other objects in the scene. This is true for all point removal methods as they simply remove points without reprojecting them back into the environment. Additionally, the added points from pattern-aware ground truth sampling are not entirely consistent with the LiDAR scanning pattern. For example, if an augmented object is placed behind another object, the augmented object should also be occluded. A future improvement to pattern-aware ground truth augmentation would be to add augmented objects that still ensure a realistic LiDAR scan.

\bibliographystyle{plain}
\bibliography{pattern_aware}


\end{document}